\documentclass{article} 
\usepackage{iclr2023_conference_tinypaper,times}

\usepackage{amsmath,amsfonts,bm}

\def\eqref#1{equation~\ref{#1}}

\def\1{\bm{1}}

\DeclareMathAlphabet{\mathsfit}{\encodingdefault}{\sfdefault}{m}{sl}
\SetMathAlphabet{\mathsfit}{bold}{\encodingdefault}{\sfdefault}{bx}{n}

\usepackage{hyperref}
\usepackage{url}
\usepackage{graphicx}

\title{NeuralCGMM: Neural Control System for Continuous Glucose Monitoring and Maintenance}

\DeclareUnicodeCharacter{2009}{\,} 

\iclrfinalcopy

\author{Azmine Toushik Wasi\\
Shahjalal University of Science and Technology,
 Bangladesh \\
\texttt{azminetoushik.wasi@gmail.com}
}
\begin{document}

\maketitle

\begin{abstract}
Precise glucose level monitoring is critical for people with diabetes to avoid serious complications. While there are several methods for continuous glucose level monitoring, research on maintenance devices is limited. To mitigate the gap, we provide a novel neural control system for continuous glucose monitoring and management that uses differential predictive control, \textit{NeuralCGMM}. Our approach, led by a sophisticated neural policy and differentiable modeling, constantly adjusts insulin supply in real-time, thereby improving glucose level optimization in the body. This end-to-end method maximizes efficiency, providing personalized care and improved health outcomes, as confirmed by empirical evidence. Code and data are available at: \url{https://github.com/azminewasi/NeuralCGMM}.
\end{abstract}

\vspace{-2mm}
\section{Introduction}
\vspace{-2mm}
Continuous glucose monitoring and maintenance play a crucial role in diabetes management, offering real-time glucose level management that empowers informed decisions on diet, medication, and lifestyle, leading to improved glycemic control and reduced risk of hypoglycemia \citep{CGM-1,CGM-2,10.2337/diaclin.34.1.25}. Yet, there remains no suitable method capable of predicting, controlling, and automating the entire process efficiently. Differentiable Predictive Control (DPC) uses a differentiable programming-based policy gradient method to train a neural network to approximate an explicit Model Predictive Control (MPC) controller without the need for supervision from an expert controller \citep{cortez2022differentiable, cortez2023robust, DPC-main-paper}. By employing differentiable system models, typically represented by ODEs, DPC adeptly adjusts to individual patient conditions, predicting and optimizing glucose levels. \\
In this work, we explore, define, and formulate the closed-loop GCM problem, and showcase a method (\textit{NeuralCGMM}) to solve it effectively with DPC. Our framework for precise glucose-level management addresses a crucial gap in current healthcare systems. Formulated as a parametric optimal control problem, it minimizes tracking errors against desired references, incorporates tailored constraints for safe insulin injection, and provides personalized care. 

\vspace{-2mm}
\section{Problem Formulation}
\vspace{-2mm}
The ultimate goal of this work is to automate insulin delivery based on continuous glucose monitoring to maintain target glucose levels in individuals with diabetes. The control objective is to minimize the deviation of the glucose levels from the target values while optimizing insulin delivery to ensure stable and controlled blood glucose concentrations. The system is represented by a differentiable model that captures the dynamics of glucose-insulin interaction. The differentiable nature allows seamless integration with DPC. The control policy, denoted as $\pi(g_k, R)$, will determine insulin delivery actions at each time step based on the continuous glucose monitoring data $g_k$ and the predictions $R = [r_k, ..., r_{k+N}]$ over a predefined prediction horizon $N$.
The parametric optimal control problem for  glucose level maintenance is formulated as follows:
\begin{equation} \label{eq:obj}
    \underset{\theta}{\text{minimize}} \sum_{i=1}^m \left( \sum_{k=1}^{N-1} Q_g \cdot \left|g^i_k - r^i_k\right| + Q_N \cdot \left|g^i_N - r^i_N\right| + Q_u \cdot \left|u^i_k - u^{i-1}_k\right| \right)
\end{equation}
subject to $g^i_{k+1} = \text{ODESolve}(f(g^i_k, u^i_k))$,  $u^i_k = \pi_{\theta}(g^i_k, R^i)$, $g^i_0 \sim \mathcal{P}_{g_0}$, $R^i \sim \mathcal{P}_R$, $u^i_k \in \mathcal{U}$, $g^i_k \in \mathcal{G}$. 

 The equations dictate the evolution of blood glucose levels, denoted as \(g^i_k\), influenced by the differentiable system model \(f(g^i_k, u^i_k)\), where \(u^i_k\) represents insulin delivery actions. The control policy \(\pi_{\theta}\) determines insulin delivery based on glucose levels \(g^i_k\) and predicted references \(R^i\). Initial glucose levels \(g^i_0\) follow a probability distribution \(\mathcal{P}_{g_0}\), and references \(R^i\) follow \(\mathcal{P}_R\). Constraints \(\mathcal{U}\) and \(\mathcal{G}\) ensure insulin delivery and glucose levels adhere to physiological limits. The formulation, incorporating weighting factors \(Q_g\), \(Q_N\), and \(Q_u\), aims to optimize insulin delivery over a prediction horizon, addressing glucose control with consideration for safety and physiological factors. The optimization process involves tuning parameters \(\theta\) using stochastic gradient descent.

\section{\textit{NeuralCGMM} Method and Experiments} \label{method}
\begin{figure}[t]
\begin{center}
\vspace{-12mm}
\includegraphics[scale=0.375]{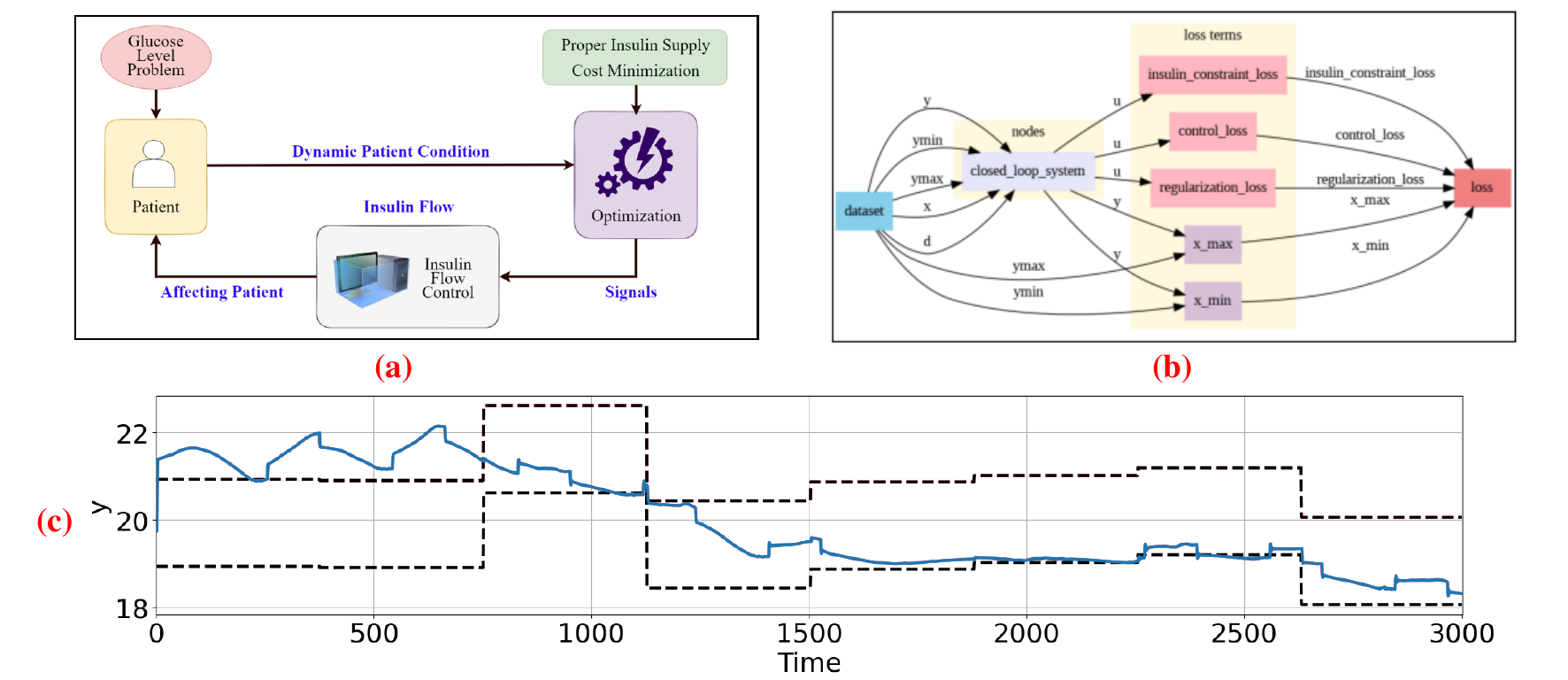}
\end{center}
\caption{(a) Continuous Glucose Monitoring and Maintenance System; (b) Model Data Flow; (c) Glucose Level Output. (Full-size figures are available in Appendix \ref{app:figures}) }\label{fig:allpic}
\vspace{-8mm}
\end{figure}

Let us consider $\mathcal{P}_{g_0}$, $\mathcal{P}_R$ and $\mathcal{P}_D$ as distributions for initial conditions, required glucose levels, and system disturbances. For model-based policy optimization, we consider a discrete-time partially observable linear state space model (SSM) \citep{SSM} that characterizes the dynamics of a patient within a medical setting as a partially observable white-box system model \citep{DRGONA202063}. The model is represented by $g_{k+1} = Ag_k + Bu_k + Ed_k ; y_k = Cg_k$. Here, \(g_k\) denotes the patient's glucose level, \(u_k\) denotes control actions governing glucose regulation, and \(d_k\) encompasses system disturbances. \(y_k\) represents the measured variable—glucose flow regulation.\\
Next, we parameterize the control policy using deep neural networks, expressed as: $u_k = \pi_{\theta}(y_k, R, D)$. Here, \(y_k\) represents the insulin flow to be controlled, \(R = \{y_{\text{min}}, y_{\text{max}}\}\) denotes the desired glucose level for the given insulin flow, and \(D\) corresponds to change in glucose level of the patient.
 With the partially observable system model and control policy in place, we formulate a differentiable closed-loop system model. 
In the closed-loop system, three penalty terms are integrated based on the objective function: control loss optimizes insulin regulation, a regularization loss deters aggressive changes in control actions to ensure patient safety, and a constraint loss limits insulin delivery rate and amount for feasibility, collectively optimizing the state variable \(g\).\\
Integrating all components, we formulate a DPC problem to be optimized comprehensively over the distribution of training scenarios, as demonstrated in Figure \ref{fig:allpic} (b). We train the model for neural control policy using stochastic gradient descent. Details are discussed in Appendix \ref{method-dep}. 

\textbf{Experiments.} 
To evaluate our model's effectiveness, we conducted training using Neuromancer \citep{Neuromancer2023} with synthetic data, detailed in the Appendix. Experimental results, illustrated in Figure \ref{fig:allpic}(c), showcase the model's adept adaptation to evolving constraints during policy updates across 3000 simulation steps, highlighting the model's responsiveness to changes in patient glucose levels and health dynamics. Details are discussed in Appendix \ref{exp-ap}.
\vspace{-2mm}
\section{Conclusion}
\vspace{-2mm}
This work introduces a significant contribution by exploring, defining, and formulating the closed-loop GCM problem through the effective application of DPC. Our innovative framework, employing DPC for precise glucose management, addresses a critical gap in healthcare systems. Utilizing a sophisticated neural policy and prioritizing resource optimization, our system detects and automatically corrects glucose irregularities, enhancing patient support in critical healthcare situations.

\clearpage
\newpage


\subsubsection*{Acknowledgements}
We extend our sincere gratitude to all the reviewers for their invaluable feedback. Additionally, we would like to express our appreciation to \hyperlink{https://ciol-sust.github.io/}{Computational Intelligence and Operations Lab - CIOL} for their support.


\subsubsection*{URM Statement}
Author Azmine Toushik Wasi meets the URM criteria of ICLR 2024 Tiny Papers Track.

\bibliography{iclr2023_conference_tinypaper}

\begin{thebibliography}{14}
\providecommand{\natexlab}[1]{#1}
\providecommand{\url}[1]{\texttt{#1}}
\expandafter\ifx\csname urlstyle\endcsname\relax
  \providecommand{\doi}[1]{doi: #1}\else
  \providecommand{\doi}{doi: \begingroup \urlstyle{rm}\Url}\fi

\bibitem[Amos et~al.(2019)Amos, Rodriguez, Sacks, Boots, and Kolter]{amos2019differentiable}
Brandon Amos, Ivan Dario~Jimenez Rodriguez, Jacob Sacks, Byron Boots, and J.~Zico Kolter.
\newblock Differentiable mpc for end-to-end planning and control, 2019.
\newblock URL \url{https://arxiv.org/pdf/1810.13400.pdf}.

\bibitem[Cortez et~al.(2022)Cortez, Drgona, Tuor, Halappanavar, and Vrabie]{cortez2022differentiable}
Wenceslao~Shaw Cortez, Jan Drgona, Aaron Tuor, Mahantesh Halappanavar, and Draguna Vrabie.
\newblock Differentiable predictive control with safety guarantees: A control barrier function approach, 2022.
\newblock URL \url{https://doi.org/10.48550/arXiv.2208.02319}.

\bibitem[Cortez et~al.(2023)Cortez, Drgona, Vrabie, and Halappanavar]{cortez2023robust}
Wenceslao~Shaw Cortez, Jan Drgona, Draguna Vrabie, and Mahantesh Halappanavar.
\newblock Robust differentiable predictive control with safety guarantees: A predictive safety filter approach, 2023.
\newblock URL \url{https://doi.org/10.48550/arXiv.2208.02319}.

\bibitem[Drgona et~al.(2023)Drgona, Tuor, Koch, Shapiro, and Vrabie]{Neuromancer2023}
Jan Drgona, Aaron Tuor, James Koch, Madelyn Shapiro, and Draguna Vrabie.
\newblock {NeuroMANCER: Neural Modules with Adaptive Nonlinear Constraints and Efficient Regularizations}.
\newblock 2023.

\bibitem[Drgoňa et~al.(2020)Drgoňa, Picard, and Helsen]{DRGONA202063}
Ján Drgoňa, Damien Picard, and Lieve Helsen.
\newblock Cloud-based implementation of white-box model predictive control for a geotabs office building: A field test demonstration.
\newblock \emph{Journal of Process Control}, 88:\penalty0 63--77, 2020.
\newblock ISSN 0959-1524.
\newblock \doi{https://doi.org/10.1016/j.jprocont.2020.02.007}.
\newblock URL \url{https://www.sciencedirect.com/science/article/pii/S0959152419306857}.

\bibitem[Drgoňa et~al.(2022)Drgoňa, Kiš, Tuor, Vrabie, and Klaučo]{DPC-main-paper}
Ján Drgoňa, Karol Kiš, Aaron Tuor, Draguna Vrabie, and Martin Klaučo.
\newblock Differentiable predictive control: Deep learning alternative to explicit model predictive control for unknown nonlinear systems.
\newblock \emph{Journal of Process Control}, 116:\penalty0 80--92, 2022.
\newblock ISSN 0959-1524.
\newblock \doi{https://doi.org/10.1016/j.jprocont.2022.06.001}.
\newblock URL \url{https://www.sciencedirect.com/science/article/pii/S0959152422000981}.

\bibitem[Lee \& Lupsa(2021)Lee and Lupsa]{CGM-1}
Grace~S Lee and Beatrice~C Lupsa.
\newblock Continuous glucose monitoring for the internist.
\newblock \emph{Med. Clin. North Am.}, 105\penalty0 (6):\penalty0 967--982, Nov 2021.
\newblock \doi{10.1016/j.mcna.2021.06.004}.
\newblock URL \url{https://doi.org/10.1016/j.mcna.2021.06.004}.

\bibitem[Ma et~al.(2022)Ma, Shao, An, Zhang, and Sun]{D2TB00749E}
Rui Ma, Ruomei Shao, Xuyao An, Qichun Zhang, and Shuqing Sun.
\newblock Recent advancements in noninvasive glucose monitoring and closed-loop management systems for diabetes.
\newblock \emph{J. Mater. Chem. B}, 10:\penalty0 5537--5555, 2022.
\newblock \doi{10.1039/D2TB00749E}.
\newblock URL \url{http://dx.doi.org/10.1039/D2TB00749E}.

\bibitem[Miller(2020)]{10.2337/cd20-0043}
Eden~M. Miller.
\newblock {Using Continuous Glucose Monitoring in Clinical Practice}.
\newblock \emph{Clinical Diabetes}, 38\penalty0 (5):\penalty0 429--438, 12 2020.
\newblock ISSN 0891-8929.
\newblock \doi{10.2337/cd20-0043}.
\newblock URL \url{https://doi.org/10.2337/cd20-0043}.

\bibitem[Oshin \& Theodorou(2023)Oshin and Theodorou]{oshin2023differentiable}
Alex Oshin and Evangelos~A. Theodorou.
\newblock Differentiable robust model predictive control, 2023.
\newblock URL \url{https://arxiv.org/pdf/2308.08426.pdf}.

\bibitem[Rangapuram et~al.(2018)Rangapuram, Seeger, Gasthaus, Stella, Wang, and Januschowski]{SSM}
Syama~Sundar Rangapuram, Matthias~W Seeger, Jan Gasthaus, Lorenzo Stella, Yuyang Wang, and Tim Januschowski.
\newblock Deep state space models for time series forecasting.
\newblock In S.~Bengio, H.~Wallach, H.~Larochelle, K.~Grauman, N.~Cesa-Bianchi, and R.~Garnett (eds.), \emph{Advances in Neural Information Processing Systems}, volume~31. Curran Associates, Inc., 2018.
\newblock URL \url{https://proceedings.neurips.cc/paper_files/paper/2018/file/5cf68969fb67aa6082363a6d4e6468e2-Paper.pdf}.

\bibitem[Schubert-Olesen et~al.(2022)Schubert-Olesen, Kröger, Siegmund, Thurm, and Halle]{ijerph191912296}
Oliver Schubert-Olesen, Jens Kröger, Thorsten Siegmund, Ulrike Thurm, and Martin Halle.
\newblock Continuous glucose monitoring and physical activity.
\newblock \emph{International Journal of Environmental Research and Public Health}, 19\penalty0 (19), 2022.
\newblock ISSN 1660-4601.
\newblock \doi{10.3390/ijerph191912296}.
\newblock URL \url{https://www.mdpi.com/1660-4601/19/19/12296}.

\bibitem[Trief et~al.(2016)Trief, Cibula, Rodriguez, Akel, and Weinstock]{10.2337/diaclin.34.1.25}
Paula~M. Trief, Donald Cibula, Elaine Rodriguez, Bridget Akel, and Ruth~S. Weinstock.
\newblock {Incorrect Insulin Administration: A Problem That Warrants Attention}.
\newblock \emph{Clinical Diabetes}, 34\penalty0 (1):\penalty0 25--33, 01 2016.
\newblock ISSN 0891-8929.
\newblock \doi{10.2337/diaclin.34.1.25}.
\newblock URL \url{https://doi.org/10.2337/diaclin.34.1.25}.

\bibitem[Yu et~al.(2023)Yu, Cho, and Lee]{CGM-2}
Jin Yu, Jae-Hyoung Cho, and Seung-Hwan Lee.
\newblock The era of continuous glucose monitoring and its expanded role in type 2 diabetes.
\newblock \emph{Journal of Diabetes Investigation}, 14\penalty0 (7):\penalty0 841--843, 2023.
\newblock \doi{https://doi.org/10.1111/jdi.14028}.
\newblock URL \url{https://onlinelibrary.wiley.com/doi/abs/10.1111/jdi.14028}.

\end{thebibliography}
\bibliographystyle{iclr2023_conference_tinypaper}

\appendix
\section{Related Works and Motivation}
\subsection{Continuous Glucose Monitoring and Maintenance}
Continuous Glucose Monitoring (CGM) technology, introduced in 1999, has emerged as a transformative tool in diabetes management. Offering real-time and predictive glycemic data, CGM systems contribute significantly to detecting trends, identifying asymptomatic events, and assessing glycemic variability. Enhanced frequency of glucose monitoring correlates with reduced hypoglycemia and increased time in range , leading to improved A1C levels. The comprehensive insights provided by CGM data analysis enable targeted treatment interventions, such as preventing hypoglycemia, optimizing glycemic control at specific times, and enhancing overall performance \citep{10.2337/cd20-0043}.

Recent advancements in CGM systems go beyond traditional blood glucose monitoring, offering robust data for effective diabetes management. \citeauthor{ijerph191912296} outlines practical recommendations for incorporating CGM into everyday physical activities, emphasizing the pivotal role of CGM in enhancing patient engagement. Additionally, \citeauthor{D2TB00749E} underscores the significance of closed-loop management systems, integrating electrochemical sensing of glucose and noninvasive monitoring technologies. These advancements not only underscore the potential of CGM in improving glucose control, especially in patients with higher initial A1C levels, but also emphasize the importance of consistent device usage for optimal benefits, setting a foundation for the future of diabetes care.

Despite notable advancements in glucose monitoring technologies, there exists a notable gap in addressing maintenance aspects, particularly the development of a system capable of autonomously administering the required insulin dosage to patients as needed. The complexity of such a system lies in its integration of both an accurate prediction policy and an optimization policy, with constraints aimed at minimizing costs and risks while simultaneously ensuring improved patient health outcomes.

\subsection{Differentiable Predictive Control} 
Differentiable Predictive Control (DPC) stands as an innovative methodology for addressing the challenges of model predictive control (MPC) in complex systems. By learning explicit neural control laws offline, DPC effectively mitigates the computational demands of online MPC. It achieves this by incorporating state and input constraints into the loss function through penalty functions and aggregating them with the MPC cost function. The resulting neural network control policy is trained offline using stochastic gradient descent, leveraging automatic differentiation of MPC problem cost functions and constraints. Demonstrating high performance with low computational resources, DPC has been successfully implemented in various applications, offering a promising avenue for enhancing control system efficiency and applicability \citep{DPC-main-paper,cortez2022differentiable,oshin2023differentiable}.

Recent advancements in DPC extend its utility by introducing it as a differentiable policy class for reinforcement learning in continuous state and action spaces. By leveraging KKT conditions and convex approximation, researchers have enabled the differentiation through MPC, allowing for end-to-end learning of the cost and dynamics of a controller. Notably, DPC exhibits superior data efficiency in comparison to generic neural networks, particularly evident in experiments involving pendulum and cartpole domains \citep{amos2019differentiable}. The approach marks a departure from traditional system identification methods, showcasing DPC's potential to outperform existing techniques and streamline learning in scenarios where expert guidance may be unavailable. 

Recognizing DPC's capability in managing intricate systems, we plan to address the maintenance challenges of current CGM systems by developing an advanced closed-loop system presented in this paper. This solution aims to enhance the effectiveness, efficiency, and optimization of continuous glucose monitoring and management using a neural network control policy. 

\section{Methodology in Depth} \label{method-dep}
\subsection{Differentiable Predictive Control}
DPC, a deep learning-driven substitute for MPC in handling unknown nonlinear systems, utilizes deep learning to grasp system dynamics from data, providing adaptability to such systems. DPC excels in managing complex real-world scenarios due to its end-to-end optimization, flexibility, and constraint support. Its ability to seamlessly handle time-varying references and constraints, along with reduced computational complexity, establishes DPC as a more efficient and versatile alternative to conventional MPC in control applications.

DPC is an algorithm rooted in model-based policy optimization. It leverages the differentiability inherent in a broad range of model representations for dynamic systems, encompassing differential equations, state-space models, and diverse neural network architectures. In the DPC framework, a differentiable closed-loop system is created, consisting of a neural control policy and a system dynamics model. The optimization process involves refining this system by utilizing parametric control objectives as intrinsic reward signals. These rewards are evaluated across a sampled distribution of the problem parameters, enabling the algorithm to iteratively improve the performance of the control policy and model \citep{DPC-main-paper}. A sample diagram is added here from the original DPC work.

The DPC problem can be formally expressed as a parametric optimal control problem, where a parametric control policy, denoted as $\pi_\theta(g(t), \xi(t))$ and parameterized by trainable weights $W$, is employed for a continuous-time dynamical system. The system is described by the differential equation:
\begin{equation} 
\frac{d g(t)}{d t} = f(g(t), {u}(t))
\end{equation}
Here, $g(t)$ represents the time-varying state, ${u}(t)$ denotes the system control inputs, and $f$ characterizes the state transition dynamics. Alternatively, in discrete-time form, obtained through methods such as ODE solvers or state-space models, the system evolves as:
\begin{equation} 
g_{k+1} = f\left(g_k, {u}_k\right)
\end{equation}
The objective is to optimize the parametric control policy to determine the control inputs, ${u}$, in order to minimize a certain cost or objective function associated with the system's behavior. The optimization is performed over the parameter space $\theta$ to enhance the policy's performance in guiding the system dynamics.
Using, this, we formulate the DPC problem as a following parametric optimal control problem presented in Equation \ref{eq:obj}.
\begin{equation} 
    \underset{\theta}{\text{minimize}}  \sum_{i=1}^m \left( \sum_{k=1}^{N-1} Q_g \cdot \left|g^i_k - r^i_k\right| + Q_N \cdot \left|g^i_N - r^i_N\right| + Q_u \cdot \left|u^i_k - u^{i-1}_k\right| \right)
\end{equation}
suject to:
\begin{align*}
& g^i_{k+1} = \text{ODESolve}\left(f(g^i_k, u^i_k)\right), \\
& u^i_k = \pi_{\theta}\left(g^i_k, R^i\right), \\
& g^i_0 \sim \mathcal{P}_{g_0}, \\
& R^i \sim \mathcal{P}_R, \\
& u^i_k \in \mathcal{U}, \\
& g^i_k \in \mathcal{G}.
\end{align*}

The primary benefit of formulating the DPC problem with differentiable closed-loop dynamics models, control objective functions, and constraints lies in the ability to employ automatic differentiation, specifically backpropagation through time. This approach facilitates the direct computation of policy gradients. By expressing the problem as a computational graph and leveraging the chain rule, we can efficiently compute the gradients of the loss function with respect to the policy parameters W. 

To obtain the gradients of the objective function with respect to the policy parameters W, we can apply the chain rule for partial differentiation. Let's denote the objective function as J:
\begin{equation}
J(\theta) = \sum_{i=1}^m \left( \sum_{k=1}^{N-1} Q_g \cdot \left|g^i_k - r^i_k\right| + Q_N \cdot \left|g^i_N - r^i_N\right| + Q_u \cdot \left|u^i_k - u^{i-1}_k\right| \right)
\end{equation}
The chain rule for partial differentiation of J with respect to the policy parameters W is expressed as:
\begin{equation}
\frac{\partial J}{\partial W} = \sum_{i=1}^m \sum_{k=1}^{N-1} \left( \frac{\partial J}{\partial g^i_k} \cdot \frac{\partial g^i_k}{\partial W} + \frac{\partial J}{\partial u^i_k} \cdot \frac{\partial u^i_k}{\partial W} \right)
\end{equation}
Here, \(\frac{\partial J}{\partial g^i_k}\) and \(\frac{\partial J}{\partial u^i_k}\) are the partial derivatives of the objective function with respect to the state and control inputs at time step k, respectively. The terms \(\frac{\partial g^i_k}{\partial W}\) and \(\frac{\partial u^i_k}{\partial W}\) represent the gradients of the state and control inputs with respect to the policy parameters W.

\subsection{White-box System Model with Control Policy}
\begin{figure}[h]
\begin{center}
\includegraphics[scale=0.9]{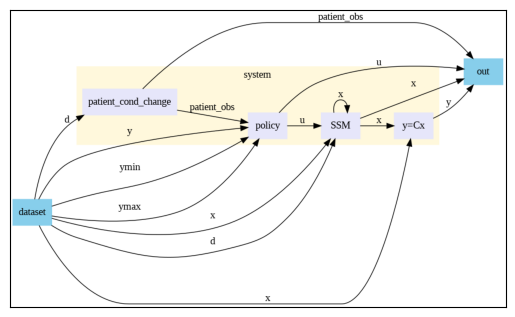}
\end{center}
\caption{Close Loop System Model. }\label{fig:CLS}
\end{figure}

As mentioned in Section \ref{method}, the control policy is parameterized using deep neural networks (MLP), specifically expressed as \(u_k = \pi_{\theta}(y_k, R, D)\). In this formulation, \(y_k\) represents the glucose level to be regulated, \(R = \{y_{\text{min}}, y_{\text{max}}\}\) denotes the desired glucose levels of the patient, and \(D\) accounts for observed disturbances, encompassing changes in the patient due to physiological factors such as eating or drinking. The neural network structure is instantiated as an MLP with bounds, characterized by an input size comprising the current glucose level, desired glucose levels, and observed disturbances. It outputs the control input \(u_k\).  The overall neural network policy, is constructed as a node in the computational graph, mapping inputs to the corresponding control output as follows in Figure \ref{fig:CLS}.

\section{Experiments and Implementation Details} \label{exp-ap}
\subsection{Parameters}
All experimental information is provided in this table \ref{tab:parameters}.
\begin{table}[h]
\centering
\caption{Parameter Settings for the System} \label{tab:parameters}
\begin{tabular}{l c}
\bf NAME & \bf VALUE \\
\hline
Desired Glucose Level, $g_{\text{min\_range}}$ & $(12., 18.)$ \\
Prediction Horizon, $n_{\text{steps}}$ & 100 \\
Samples, $n_{\text{samples}}$ & 2000  \\
$\text{Batch Size}$ & 64 \\
\text{Control Model} & MLP \\
\text{MLP Specification} & \text{2 hidden layers with 32 units in each} \\
\text{MLP Activation} & \text{GELU} \\
\text{Control Loss Hyperparameter} & $Q_g = 0.01$ \\
\text{Regularization Loss Hyperparameter} & $Q_N = 0.1$ \\
\text{Constraint Loss Hyperparameter} & $Q_u = 0.02$ \\
\text{Trainer Parameters: Epochs} & $200$ \\
\text{Trainer Parameters: Warmup} & 50 \\
\text{Trainer Parameters: Optimizer} & \text{AdamW} \\
\text{Trainer Parameters: Learning Rate} & $0.001$ \\
\text{Trainer Parameters: Early Stopping} & \text{No update in} 5 \text{steps} \\
\end{tabular}
\end{table}

\subsection{Dataset Generation} \label{dataset-gen}
The data generation process involves creating training and development datasets for a dynamic system. For each scenario, initial conditions are sampled from the system, and a prediction horizon of 100 steps is defined. The optimal is sampled from a uniform distribution between 18 and 20. Disturbance trajectories are generated using the system's simulation model. The data is organized into batches for efficient training, with 64 sampled scenarios for each batch. Finally, dataloader instances are created for both training and development datasets, facilitating the training of the DPC algorithm on the dynamic system.

\subsection{Test}
For testing, we randomly generated a test set of 3000 datapoints as described in Section \ref{dataset-gen}, and generated predictions using the model. 

\begin{figure}[h]
\begin{center}
\includegraphics[scale=0.275]{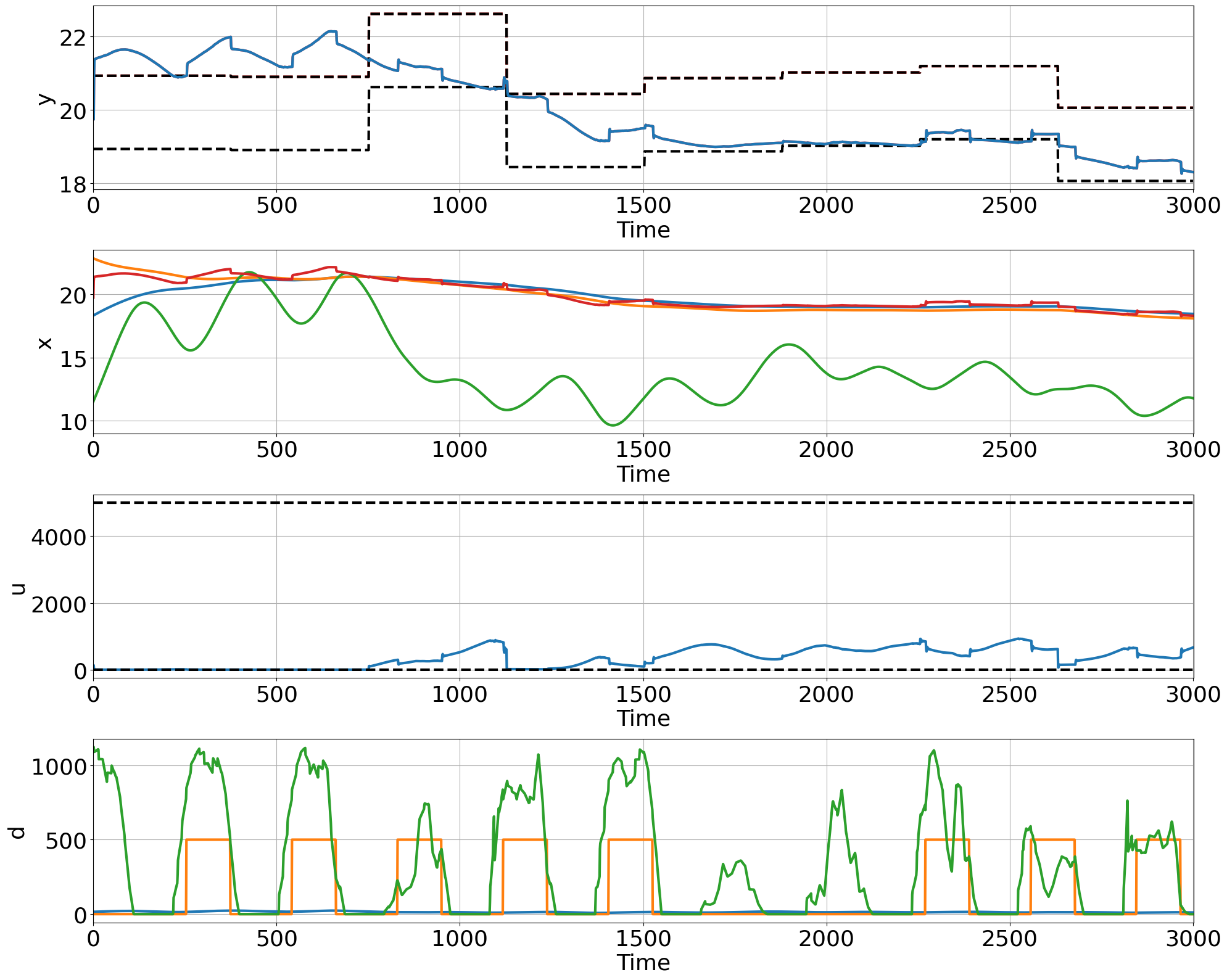}
\end{center}
\caption{Test Output Visualization. }\label{fig:Test-Result}
\end{figure}

In Figure \ref{fig:Test-Result}, we can see the output and model parameters for different values in the synthetic test set. We can see in the $y$ chart that the control policy is trying to keep the value within limits, though there are several changes in the system. $u$ is also within the constraints. Also, the $d$ chart shows that the model is capturing the disturbances nicely.

\textit{ The best model and test data samples, along with the codes, are included in the supplementary material for result reproduction purposes.}

\newpage

\subsection{Figures} \label{app:figures}
Here we include a zoomed version of the two charts in Figure \ref{fig:allpic}. The another is already added in Figure \ref{fig:Test-Result} chart 1.

\begin{figure}[h]
\begin{center}
\includegraphics[scale=0.225]{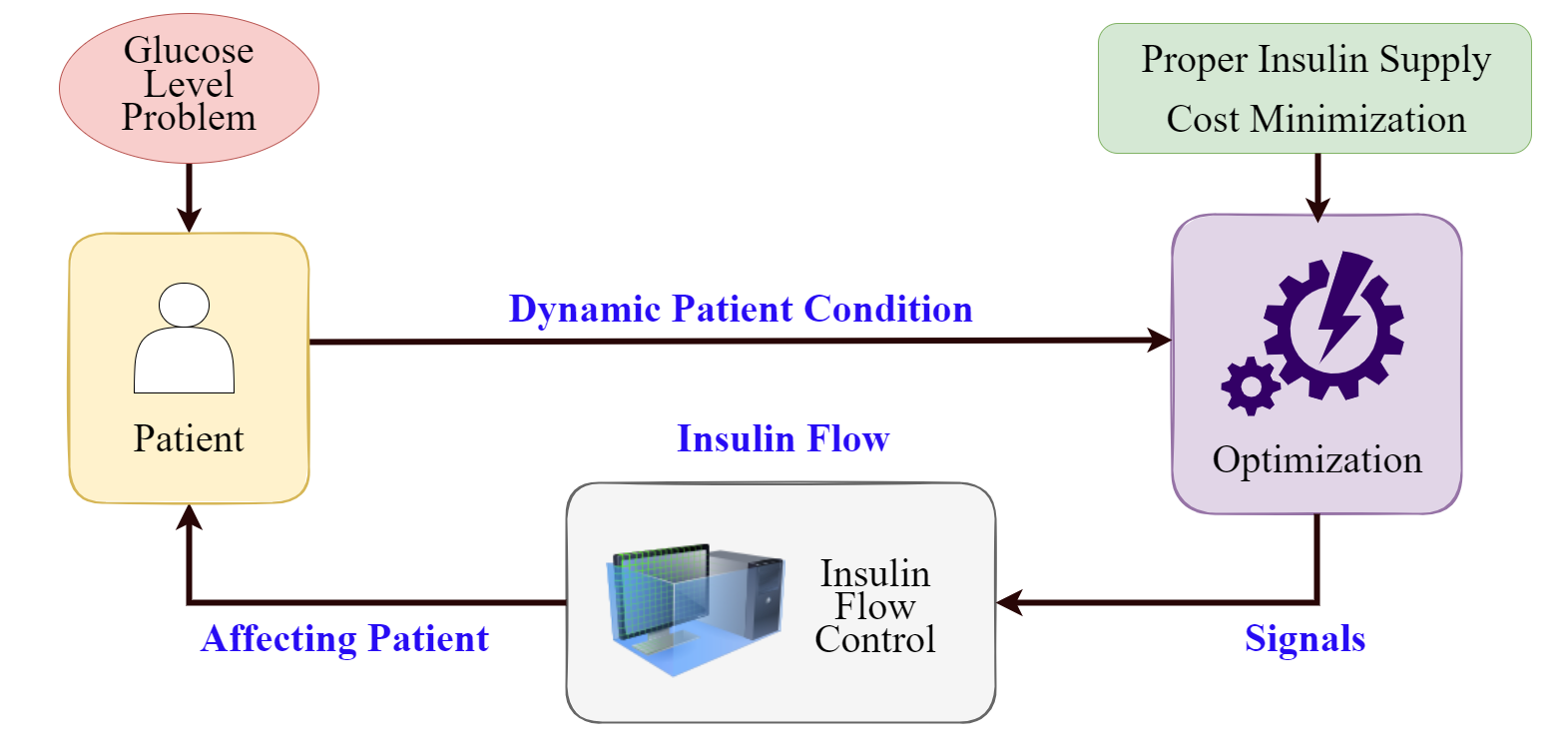}
\end{center}
\caption{Continuous Glucose Monitoring and Maintenance System }\label{fig:diagram}
\end{figure}

\begin{figure}[h]
\begin{center}
\includegraphics[scale=0.95]{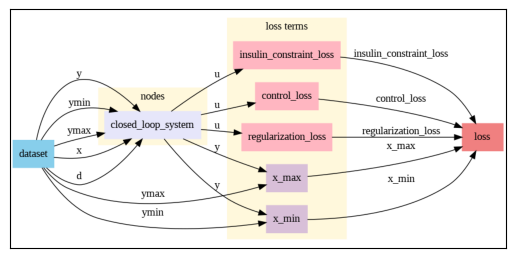}
\end{center}
\caption{Model Data Flow for Continuous Glucose Monitoring and Maintenance System  }\label{fig:problem}
\end{figure}

\section{Discussion}
This study establishes a theoretical foundation for monitoring and maintaining glucose levels through differential predictive control. We feel that it is a scalable procedure that requires in-depth research and development with clinical objectives in order to develop a practical healthcare solution. 

\textbf{Study Limitations due to Synthetic Data.} We acknowledge that synthetically generated data has limitations and might not accurately represent the intricacies of real-world situations. It is important to point out that this relates to a clinical procedure in which errors could result in patient deaths. As the research on active glucose level maintenance is still limited, we are unable to test it with real-world data.
Using synthetically generated data, this work provides a foundational framework for further exploration and clinical studies. 

\textbf{Scalability.}
Usual healthcare devices face scalability challenges in diverse clinical settings due to varying patient conditions and limitations. However, DPC offers a promising solution by dynamically adapting to patient dynamics, mitigating the impact of clinical variations. Its ability to handle personalized constraints and optimize insulin delivery makes it a scalable and adaptable approach for addressing diverse clinical limitations in our parametric optimal control framework.
Effective implementation of this approach requires extensive research, rigorous testing, meticulous calibration, and thorough clinical trials to setup and analyze different parameters and factors, ultimately leading to the development of a practical product that addresses glucose level management effectively.

\textbf{Limitations.}
There are certain shortcomings and problems that may arise that should be considered. 
The extensive calibration required for the practical application of our approach can be a practical obstacle in real-world healthcare settings, as can the complexity of conducting rigorous clinical testing during the development stage. These tasks can be costly and time-consuming, which may hinder the broad adoption of our approach.
Lastly, unexpected external factors and technological limitations may affect the precision and dependability of glucose level predictions, which we believe is a problem for almost all devices.

Through this work, we hope to pave the way for future research initiatives, ultimately contributing to the development of a refined device that holds the potential to positively impact the lives of millions worldwide.

\end{document}